\documentclass{article}

     \PassOptionsToPackage{numbers, compress}{natbib}

     \usepackage[final]{neurips_2020}

\usepackage[utf8]{inputenc} 
\usepackage[T1]{fontenc}    
\usepackage{hyperref}       
\usepackage{url}            
\usepackage{booktabs}       
\usepackage{amsfonts}       
\usepackage{nicefrac}       
\usepackage{microtype}      

\usepackage{graphicx}
\usepackage{amsmath}
\usepackage{subfigure}
\usepackage{multirow}
\usepackage[ruled,vlined]{algorithm2e}
\usepackage{wrapfig}
\usepackage{enumitem}

\title{A Meta-Learning Approach for Graph Representation Learning in Multi-Task Settings}

\author{%
  Davide Buffelli\\
  Department of Information Engineering\\
  University of Padova\\
  Padova, Italy \\
  \texttt{davide.buffelli@unipd.it} \\
  \And
  Fabio Vandin\\
  Department of Information Engineering\\
  University of Padova\\
  Padova, Italy \\
  \texttt{fabio.vandin@unipd.it} \\
}

\begin{document}

\maketitle

\begin{abstract}
Graph Neural Networks (GNNs) are a framework for \textit{graph representation learning}, where a model learns to generate low dimensional node embeddings that encapsulate structural and feature-related information.
GNNs are usually trained in an end-to-end fashion, leading to highly specialized node embeddings. 
However, generating node embeddings that can be used to perform multiple tasks (with performance comparable to single-task models) is an open problem.
We propose a novel meta-learning strategy capable of producing \textit{multi-task} node embeddings. Our method avoids the difficulties arising when learning to perform multiple tasks \emph{concurrently} by, instead, learning to quickly (i.e. with a few steps of gradient descent) adapt to multiple tasks \textit{singularly}. 
We show that the embeddings produced by our method can be used to perform multiple tasks with comparable or higher performance than classically trained models. 
Our method is model-agnostic and task-agnostic, thus applicable to a wide variety of multi-task domains.
\end{abstract}

\section{Introduction}\label{intro}
\begin{wrapfigure}{R}{0.58\textwidth}
\centering
  \subfigure[]{\includegraphics[width=0.18\textwidth]{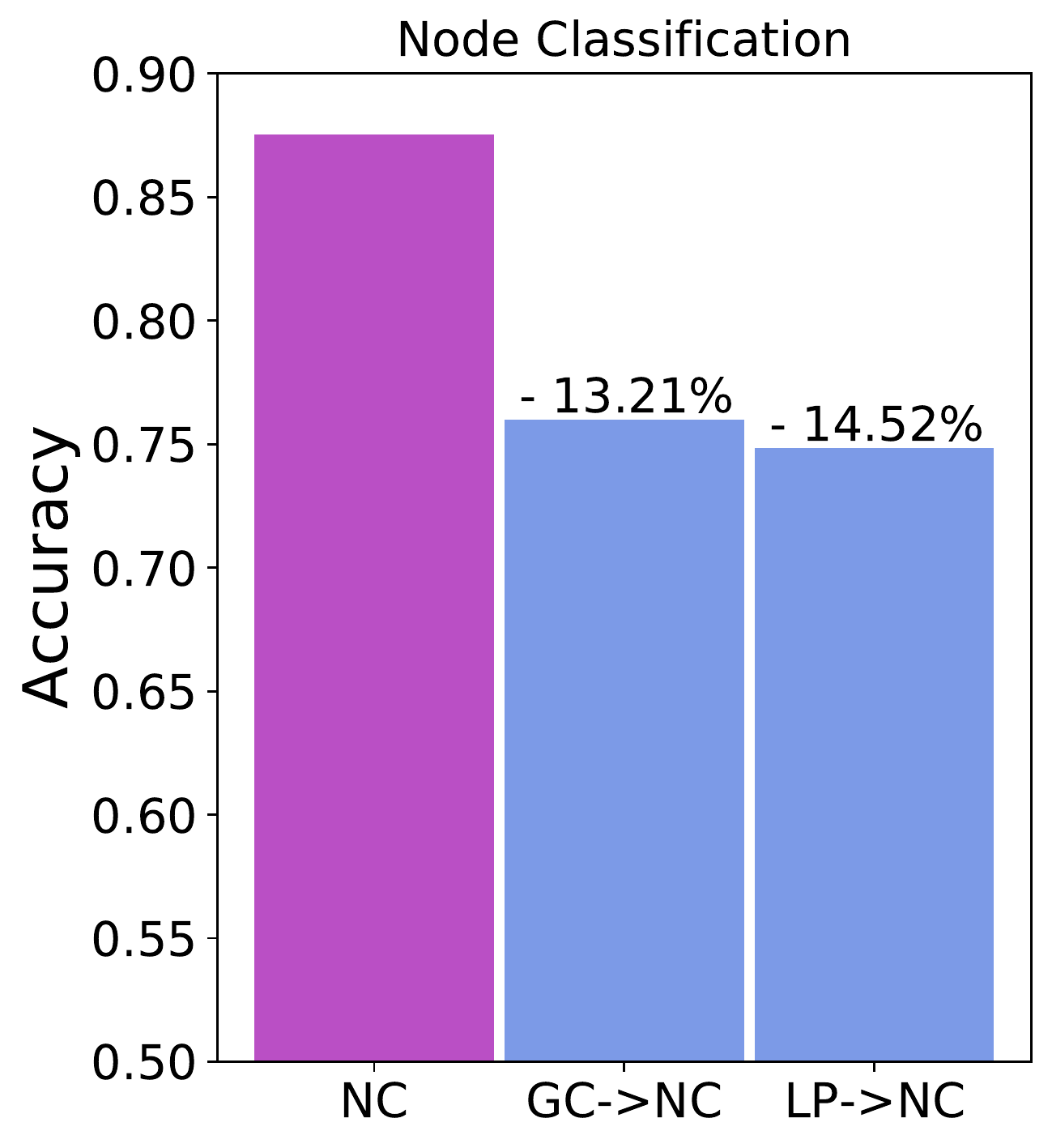}}
  \subfigure[]{\includegraphics[width=0.18\textwidth]{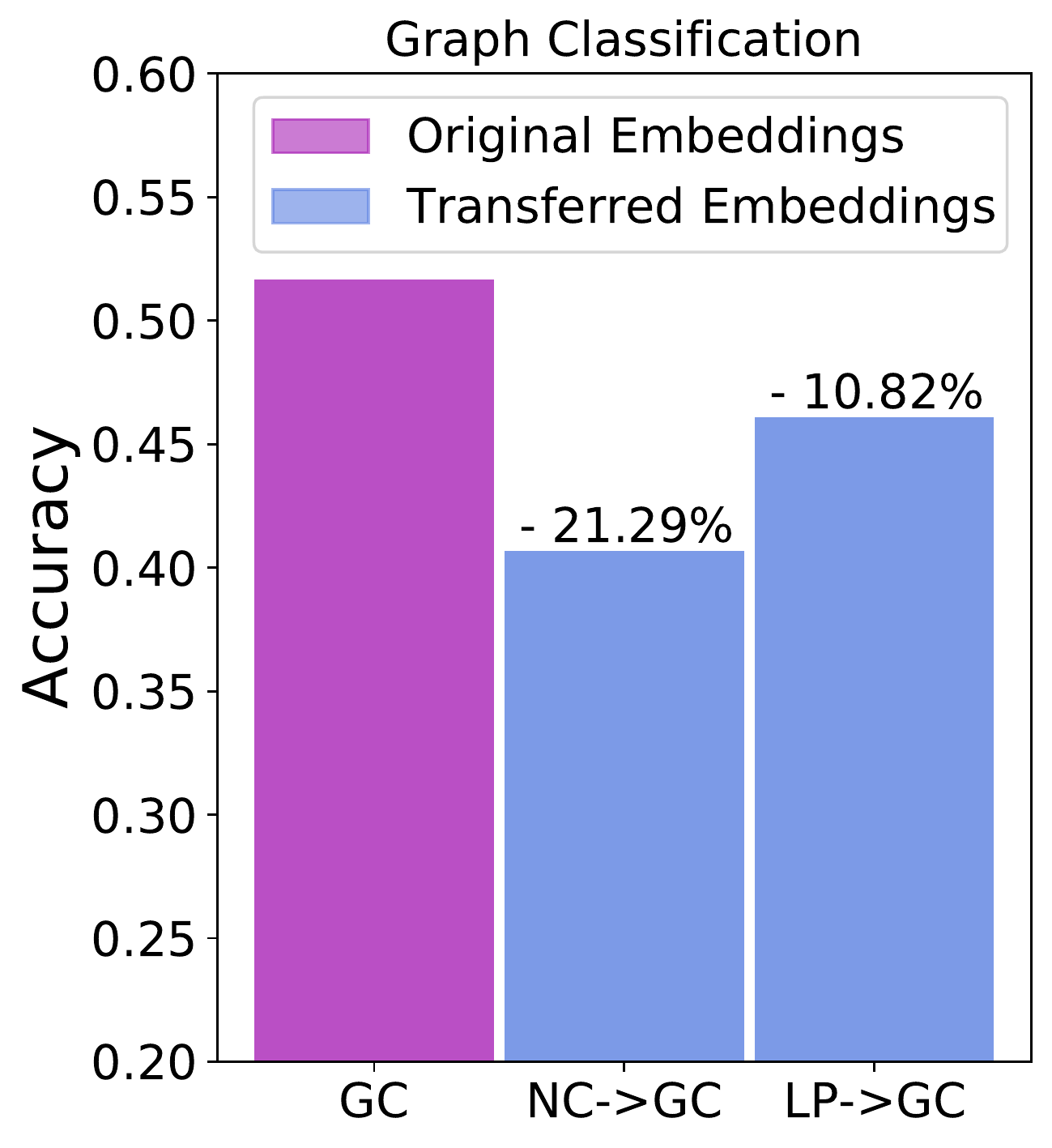}}
  \subfigure[]{\includegraphics[width=0.18\textwidth]{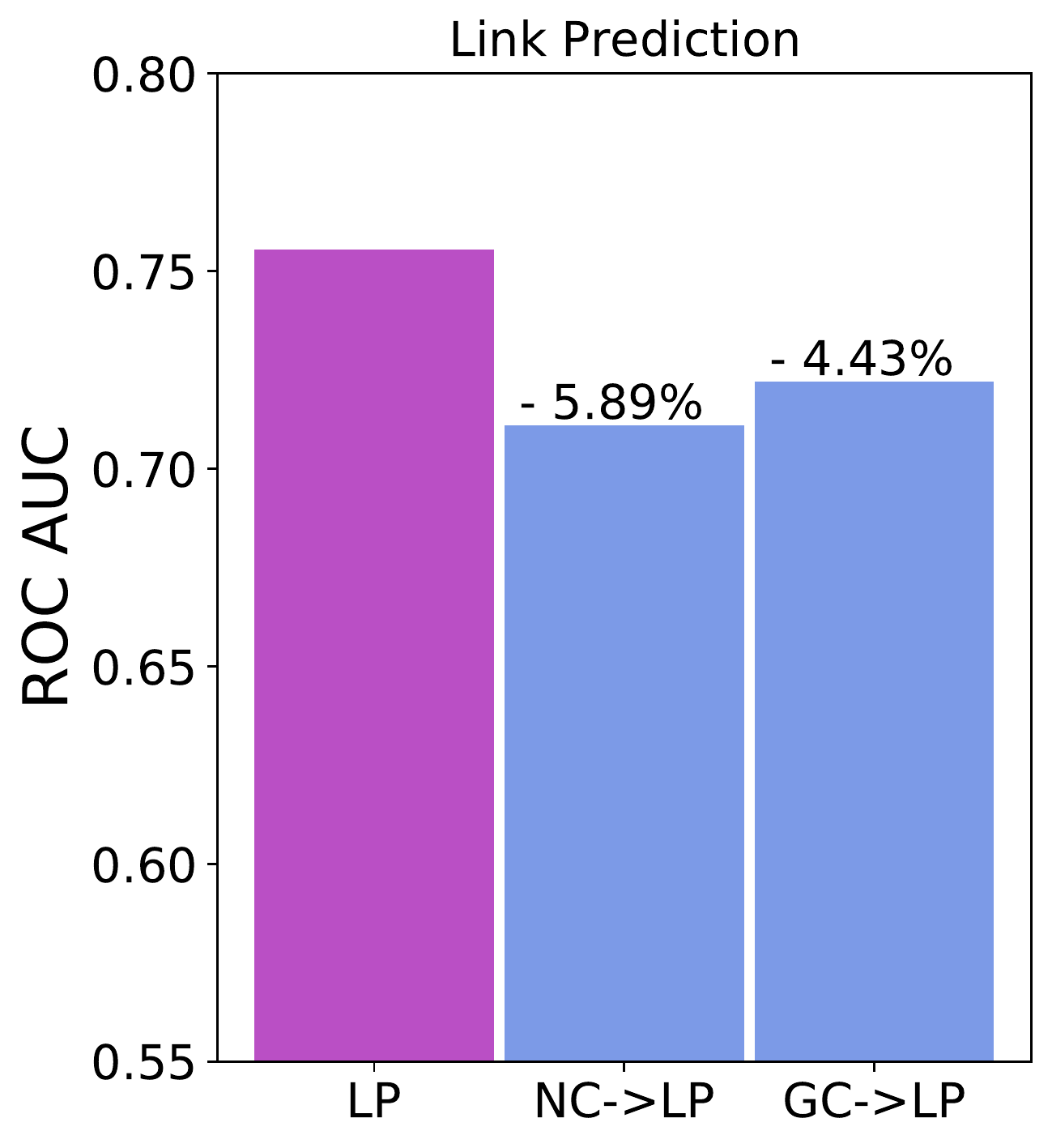}}
  \caption{Performance drop when transferring node embeddings between (a) Node Classification (NC), (b) Graph Classification (GC), and (c) Link Prediction (LP) on the ENZYMES dataset. ``\textit{x -\textgreater y}'' indicates that the embeddings obtained from a model trained on task \textit{x} are used for task \textit{y}.}
  \label{fig:transfer_embeddings}
\end{wrapfigure}
Graph Neural Networks (GNNs) are deep learning models that operate on graph structured data obtaining great empirical performance, and are a very active area of research. Three tasks in particular have received the most attention: graph classification, node classification, and link prediction.
GNNs are centered around the concept of \textit{node representation learning}, and typically follow the same architectural pattern with an \textit{encoder-decoder} structure \citep{hamilton2017representation,Chami2020MachineLO,Wu_2020}. The encoder produces node embeddings (low-dimensional vectors capturing structural and feature-related information about each node), while the decoder uses the embeddings to carry out the desired downstream task. The model is then trained in an end-to-end manner, giving rise to highly specialized node embeddings. 
In fact, taking the embeddings from a trained GNN, and using them to train a decoder for a different task, leads to substantial performance loss (see Figure \ref{fig:transfer_embeddings}).

The low transferability of node embeddings requires the use of task-specific encoders and decoders. However, many practical machine learning applications operate in resource-constrained environments where being able to share parameters between tasks is of great importance. 
Learning models that perform multiple tasks is known as \textit{Multi-Task Learning} (MTL), and is an open area of research \citep{v2020revisiting}.

We show that training a multi-head model with the classical procedure, i.e. by performing multiple tasks concurrently on each graph, and updating the parameters with some form of gradient descent to minimize the sum of the single-task losses, can lead to performance loss with respect to single-task models.
We then propose a novel optimization-based meta-learning \citep{finn2017} procedure that can generate node embeddings that generalize across tasks. 
Our meta-learning procedure does not aim 
at a setting of the parameters that can perform multiple tasks concurrently (like a classical method would do), or to a setting that allows fast multi-task adaptation (like traditional meta-learning), but to a setting that can \textbf{easily be adapted to perform each of the tasks singularly}. In fact, our procedure aims at a setting of the parameters where a few steps of gradient descent on a given task, can lead to good performance on that task, hence removing the burden of learning to solve multiple tasks \textit{concurrently}.

We summarize our contributions as follows:
\begin{itemize}
\item We propose a novel meta-learning strategy for multi-task representation learning. We apply it on graph MTL, and show that a GNN trained with our method produces higher quality node embeddings with respect to classical training procedures. Our method is \textit{model-agnostic} and \textit{task-agnostic}, thus easily applicable to a wide range of multi-task domains.
\item To the best of our knowledge, we are the first to propose a GNN model generating a \textit{single} set of node embeddings that can be used to perform the three most common graph-related tasks. 
In fact, our embeddings lead to comparable or higher performance with respect to single-task models even when used as input to a simple linear classifier.
\item We show that the episodic training strategy in our meta-learning procedure leads to better node embeddings even for single-task models. We believe this finding provides interesting directions for future work on connections between meta-learning and representation learning.
\end{itemize}

\section{Related Work}\label{rel_work}
GNNs, MTL, and meta-learning are very active areas of research. We highlight works that are at the intersections of these subjects, and point the interested reader to comprehensive reviews of each field. 

\textbf{Graph Neural Networks.}
GNNs have a long history \citep{4700287}, but in the past few years the field has grown exponentially. 
Seminal works include ChebNet \citep{cnn_graph}, GCN \citep{kipf2017semi}, GAT \citep{velickovic2018graph}, and GIN \citep{xu2018how}. 
For a thorough review of the field we refer the reader to \citet{Chami2020MachineLO} and \citet{Wu_2020}. 

\textbf{Multi-Task Learning.}
Works at the intersection of MTL and GNNs have focused on multi-head architectures for several applications \citep{Montanari_2019,holtz4multi,xu2018how,Avelar_2019,li-ji-2019-syntax}, but no \textit{single} model has been proposed for the three most common tasks on graphs. Other works use GNNs as a tool for MTL: \citet{Liu2019LearningMC} use GNNs to allow communication between tasks, while \citet{10.5555/3327345.3327479} use GNNs to estimate the test error of a MTL model.
For an exhaustive review of deep MTL we refer to \citet{v2020revisiting}.

\textbf{Meta-Learning.}
Meta-Learning has attracted considerable attention (see the review by \citet{hospedales2020metalearning}), specially in the area of \textit{few-shot learning}. 
Some works use GNNs directly for few-shot learning \citep{garcia2017few}, others as a tool for enhancing meta-learning \citep{liu2019GPN,10.1145/3394486.3403230}, and others use meta-learning to train GNNs in few-shot learning scenarios for graph-related problems \citep{10.1145/3357384.3358106,yeaofewshot2020,Kim2019EdgeLabelingGN,Kim2019EdgeLabelingGN,Alet2019NeuralRI,bose2019meta,nguyen2020}, 
Other works combining meta-learning and GNNs involve adversarial attacks \citep{zugner_adversarial_2019} and active learning \citep{madhawa2020}.

\section{Preliminaries}

\subsection{Graph Neural Networks}
Many GNNs follow the \textit{message-passing} paradigm \citep{10.5555/3305381.3305512}. Let us represent a graph $\mathcal{G} = (\mathbf{A}, \mathbf{X})$ with an adjacency matrix $\mathbf{A} \in \{0,1\}^{n \times n}$, and a node feature matrix $\mathbf{X} \in \mathbb{R}^{n \times d}$, where the $v$-th row $\mathbf{X}_v$ represents the $d$ dimensional feature vector of node $v$. Let $\mathbf{H}^{(\ell)} \in \mathbb{R}^{n \times d^{\prime}}$ be the node representation matrix at layer $\ell$. A message passing layer updates the representation of every node $v$ as follows:
\[
\text{\textit{msg}}^{(\ell)}_{v}  = \text{AGGREGATE}(\{ \mathbf{H}_{u}^{(\ell)} \text{ } \forall u \in \mathcal{N}_v \}) , \quad 
\mathbf{H}_{v}^{(\ell+1)} = \text{UPDATE}(\mathbf{H}_{v}^{(\ell)}, \text{\textit{msg}}^{(\ell)}_{v})   
\]
where $\mathbf{H}^{(0)} = \mathbf{X}$, $\mathcal{N}_v$ is the set of neighbours of node $v$, $\text{AGGREGATE}$ is a permutation invariant function, and $\text{UPDATE}$ is usually a neural network. After $L$ message-passing layers, the final node embeddings $\mathbf{H}^{(L)}$ are used to perform a given task, and the network is trained end-to-end. 

\subsection{Model-Agnostic Meta-Learning and ANIL}\label{maml_anil}
MAML (Model-Agnostic Meta-Learning)\citep{finn2017} is an optimization-based meta-learning strategy. Let $f_\theta$ be a deep learning model, where $\theta$ are its parameters. Let $p(\mathcal{E})$ be a distribution over episodes\footnote{The meta-learning literature usually derives episodes from \textit{tasks} (i.e. tuples containing a dataset and a loss function). We focus on episodes to avoid using the term \textit{task} for both a MTL task, and a meta-learning task.}. An episode $\mathcal{E}_i \sim p(\mathcal{E})$ is defined as a tuple containing a \textit{loss function} $\mathcal{L}_{\mathcal{E}_i}$, a \textit{support set} $ \mathcal{S}_{\mathcal{E}_i}$, and a \textit{target set} $\mathcal{T}_{\mathcal{E}_i}$: $\mathcal{E}_i = (\mathcal{L}_{\mathcal{E}_i}(\cdot), \mathcal{S}_{\mathcal{E}_i}, \mathcal{T}_{\mathcal{E}_i})$ (support and target sets are sets of labelled examples). MAML's goal is to find a value of $\theta$ that can quickly, i.e. in a few steps of gradient descent, be adapted to new episodes. This is done with a nested loop optimization procedure: an \textit{inner loop} adapts the parameters to the support set of an episode by performing some steps of gradient descent, and an \textit{outer loop} updates the initial parameters to allow fast adaptation. Formally, let $\theta^{\prime}_{i} (t)$ be the parameters after $t$ adaptation steps on the support set of episode $\mathcal{E}_i$, then the computations in the inner loop are
\[ \theta^{\prime}_{i} (t) = \theta^{\prime}_{i} (t-1) - \alpha \nabla_{\theta^{\prime}_{i} (t-1)} \mathcal{L}_{\mathcal{E}_i}(f_{\theta^{\prime}_{i} (t-1)}, \mathcal{S}_{\mathcal{E}_i}) , \text{ with } \theta^{\prime}_{i} (0) = \theta \]
where $\mathcal{L}(f_{\theta^{\prime}_{i} (t-1)}, \mathcal{S}_{\mathcal{E}_i})$ indicates the loss over the support set $\mathcal{S}_{\mathcal{E}_i}$ of the model with parameters $\theta^{\prime}_{i} (t-1)$, and $\alpha$ is the learning rate. The \textit{meta-objective} that the outer loop tries to minimize is defined as
$ \mathcal{L}_{\text{\textit{meta}}} = \sum_{\mathcal{E}_{i} \sim p(\mathcal{E})} \mathcal{L}_{\mathcal{E}_i}(f_{\theta^{\prime}_{i} (t)}, \mathcal{T}_{\mathcal{E}_i}) $,
which leads to the following parameter update\footnote{We limit ourself to one step of gradient descent for clarity, but any optimization strategy could be used.}
\[ \theta = \theta - \beta \nabla_{\theta}\mathcal{L}_{\text{\textit{meta}}} = \theta - \beta \nabla_{\theta} \sum_{\mathcal{E}_{i} \sim p(\mathcal{E})} \mathcal{L}_{\mathcal{E}_i}(f_{\theta^{\prime}_{i} (t)}, \mathcal{T}_{\mathcal{E}_i}). \]
\citet{raghu2019rapid} showed that feature reuse is the dominant factor in MAML: in the adaptation loop, only the last layer(s) in the network are updated, while the first layer(s) remain almost unchanged. The authors then propose ANIL (Almost No Inner Loop) where they split the parameters in two sets: one that is used for adaptation in the inner loop, and one that is only updated in the outer loop. This simplification leads to computational improvements while maintaining performance.

\section{Our Method}
Our novel representation learning technique, based on meta-learning, is built on three insights: 

\textbf{\textit{(i)} optimization-based meta-learning is implicitly learning robust representations.} The findings by \citet{raghu2019rapid} suggest that in a model trained with MAML, the first layer(s) learn features that are reusable across episodes, while the last layer(s) are set up for fast adaptation. MAML is then \textit{implicitly} focusing on learning reusable representations that generalize across episodes. 

\textbf{\textit{(ii)} meta-learning episodes can be designed to encourage generalization.} If we design support and target set to mimic the training and validation sets of a classical training procedure, then the meta-learning procedure is effectively optimizing for generalization.

\textbf{\textit{(iii)} meta-learning can learn to quickly adapt to multiple tasks \textit{singularly}, without having to learn to solve multiple tasks \textit{concurrently}.} 
We design the meta-learning procedure so that, for each considered task, the inner loop adapts the parameters to a task-specific support set, and tests the adaptation on a task-specific target set. The outer loop then updates the parameters to allow fast \textit{multiple} \textbf{single-task adaptation}.
This strategy is searching for a parameter setting that can be easily adapted for good single-task performance, without learning to solve multiple tasks concurrently. (See Appendix \ref{appendix_diff} for a comparison with classical training and meta-learning strategies.)

Based on \textbf{\textit{(ii})} and \textbf{\textit{(iii})}, we develop a novel meta-learning procedure where the inner loop adapts to multiple tasks \textit{singularly}, each time with the goal of single-task generalization. Using an encoder-decoder architecture, and episodes that involve adapting to multiple tasks, \textbf{\textit{(i})} suggests that this procedure leads to an encoder that learns features that are reusable across episodes (and hence tasks). 

\textbf{\textit{Intuition.}} Training multi-task models is challenging, as tasks may negatively interfere with each other \citep{st2019tasks}.
We design a meta-learning procedure where the learner does not have to find a configuration of the parameters that \textit{concurrently} performs all tasks, but a configuration that can \textbf{easily be adapted to perform each of the tasks singularly}. Finally, leveraging the robust representation learning that happens with MAML and ANIL, we can extract an encoder generating node representations that generalize across tasks.

We now formally present our novel meta-learning procedure in three steps:
\textbf{(1) Episode Design:} how is a an episode composed, \textbf{(2) Model Architecture Design:} what is the architecture of our model, \textbf{(3) Meta-Training Design:} how, and which, parameters are adapted/updated.

\begin{figure}[t]
\centering
  \subfigure[]{\includegraphics[width=0.26\textwidth]{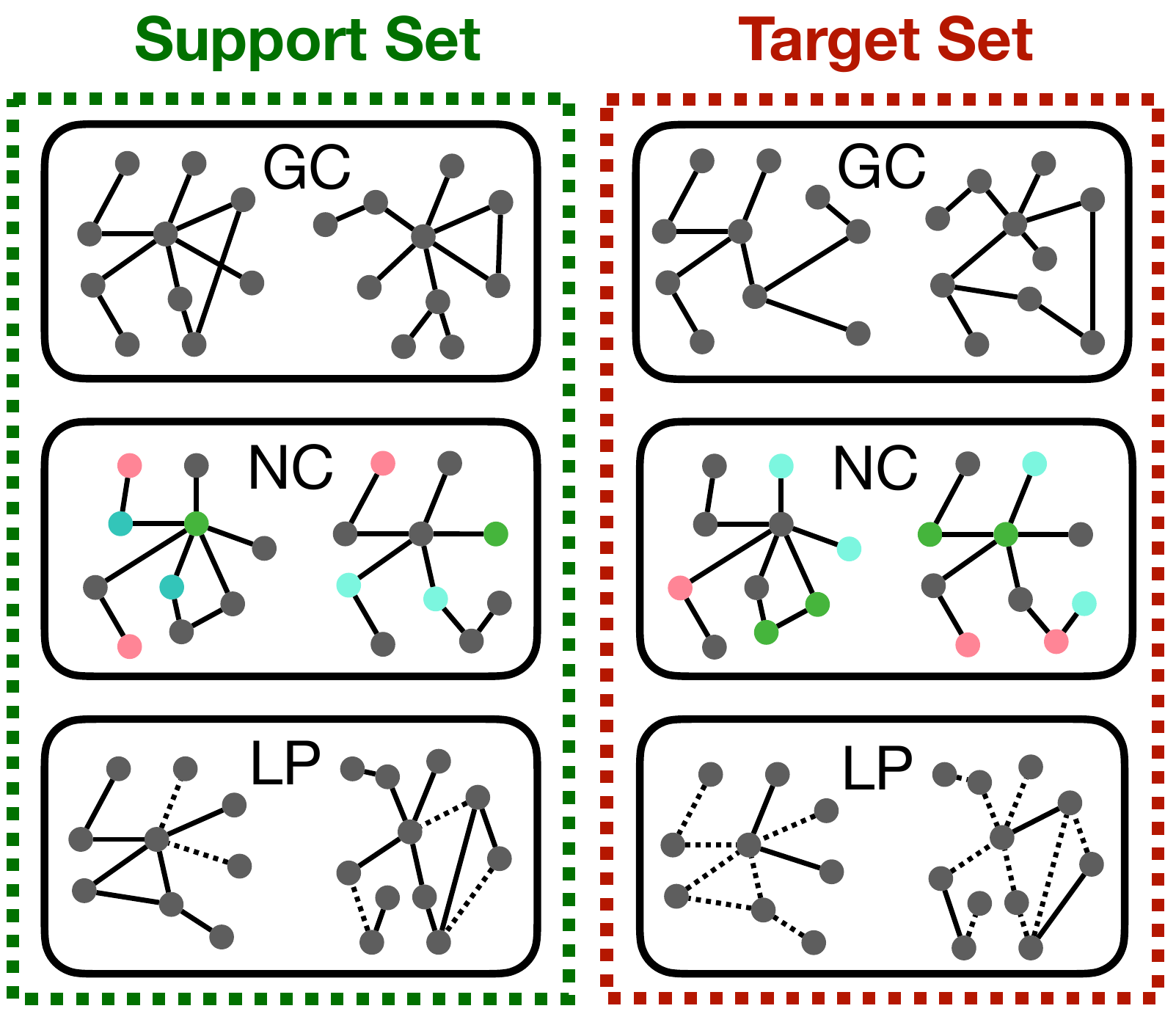}}\hspace{\fill} \vline \hspace{\fill}
  \subfigure[]{\includegraphics[height=1.25in]{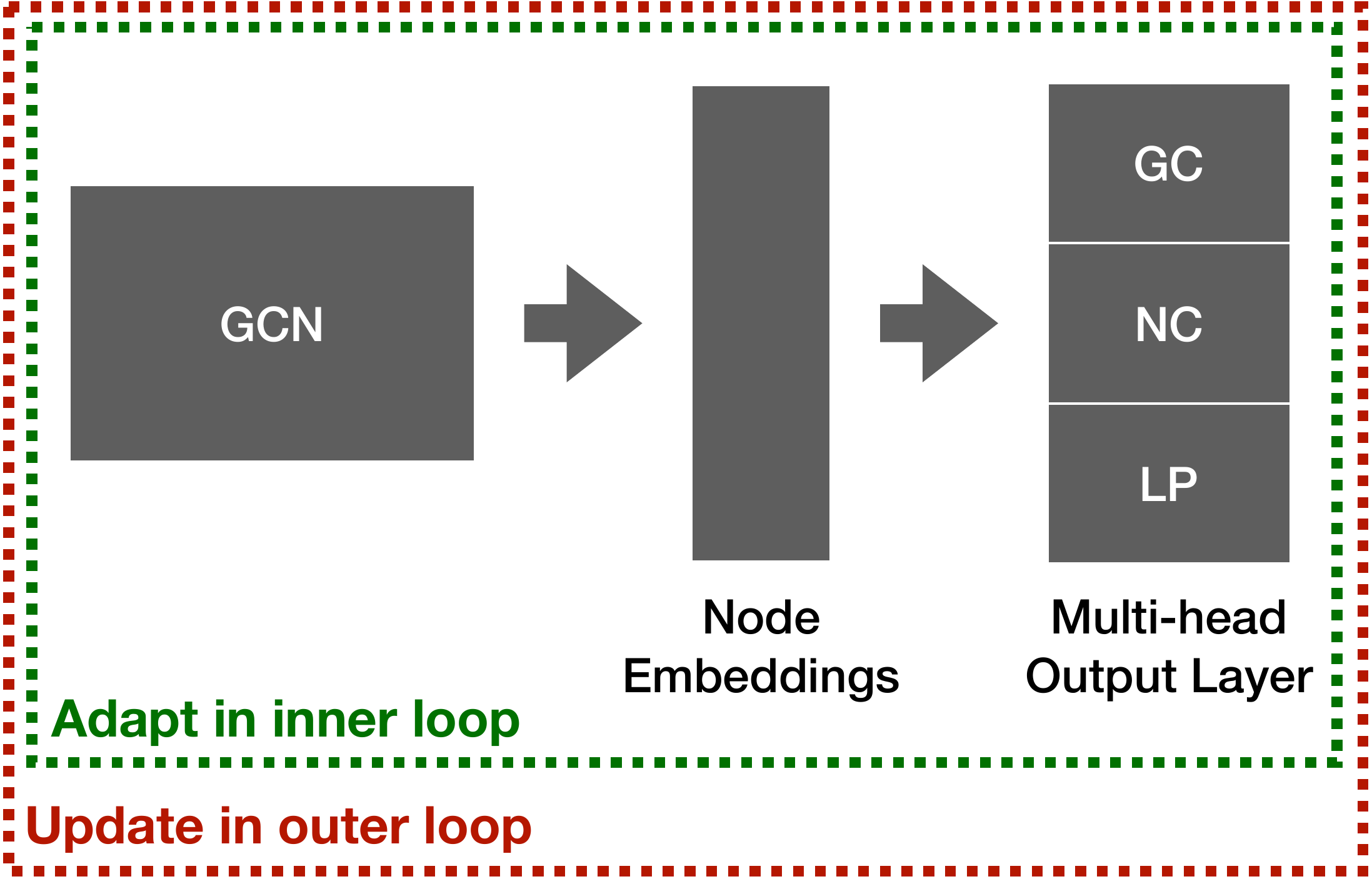}}\hspace{\fill}
  \subfigure[]{\includegraphics[height=1.25in]{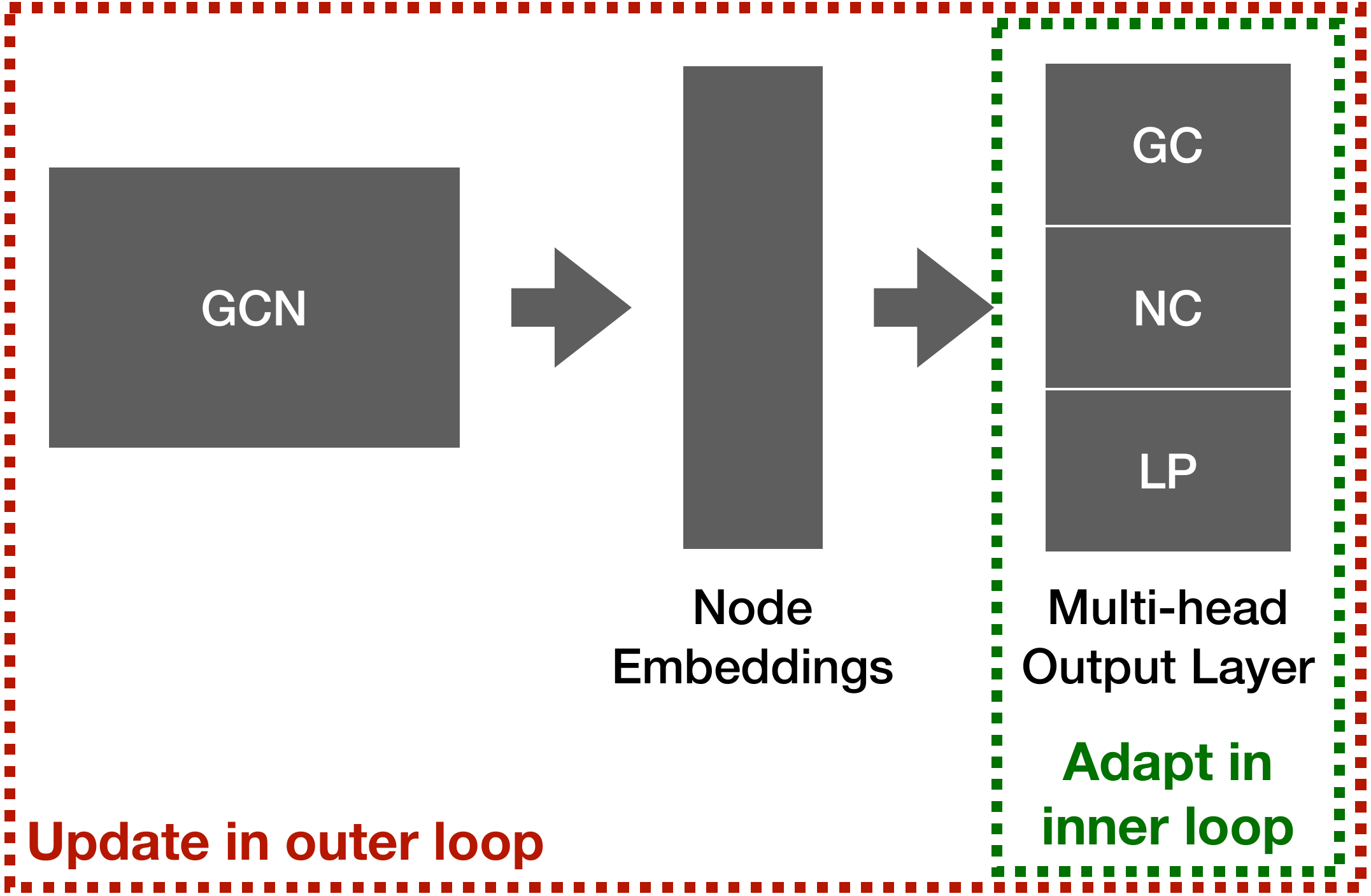}}
  \caption{(a) \textit{Multi-task episode}: for each task, support and target sets mimic training and validation sets. (b) iSAME: both backbone and task-specific output layers are adapted (one at a time) in the inner loop. (c) eSAME: only task-specific output layers are adapted (one at a time) in the inner loop.}
  \label{fig:models}
\end{figure}
\subsection{Episode Design} 
In our case, an episode becomes a \textit{multi-task episode} (Figure \ref{fig:models} (a)). Let us consider the case where the tasks are graph classification (GC), node classification (NC), and link prediction (LP). We define a \textit{multi-task episode} $\mathcal{E}^{(m)}_{i} \sim p(\mathcal{E}^{(m)})$ as a tuple
$\mathcal{E}^{(m)}_{i} = ( \mathcal{L}^{(m)}_{\mathcal{E}_{i}},  \mathcal{S}^{(m)}_{\mathcal{E}_{i}},   \mathcal{T}^{(m)}_{\mathcal{E}_{i}} )$, with
\begin{align*}
\mathcal{L}^{(m)}_{\mathcal{E}_{i}} &= \lambda^{(GC)} \mathcal{L}^{(\text{GC})}_{\mathcal{E}_{i}} + \lambda^{(NC)} \mathcal{L}^{(\text{NC})}_{\mathcal{E}_{i}} + \lambda^{(LP)} \mathcal{L}^{(\text{LP})}_{\mathcal{E}_{i}} \\
\mathcal{S}^{(m)}_{\mathcal{E}_{i}} &= \{ \mathcal{S}^{(\text{GC})}_{\mathcal{E}_i}, \mathcal{S}^{(\text{NC})}_{\mathcal{E}_i}, \mathcal{S}^{(\text{LP})}_{\mathcal{E}_i} \}, \quad
\mathcal{T}^{(m)}_{\mathcal{E}_{i}} = \{ \mathcal{T}^{(\text{GC})}_{\mathcal{E}_i}, \mathcal{T}^{(\text{NC})}_{\mathcal{E}_i}, \mathcal{T}^{(\text{LP})}_{\mathcal{E}_i} \}
\end{align*}
where $\lambda^{(\cdot)}$ are balancing coefficients.
The meta-objective of our method then becomes:
\[ \mathcal{L}^{(m)}_{\text{\textit{meta}}} = \sum_{\mathcal{E}^{(m)}_{i} \sim p(\mathcal{E}^{(m)})} \lambda^{(GC)} \mathcal{L}^{(\text{GC})}_{\mathcal{E}_{i}} + \lambda^{(NC)} \mathcal{L}^{(\text{NC})}_{\mathcal{E}_{i}} + \lambda^{(LP)} \mathcal{L}^{(\text{LP})}_{\mathcal{E}_{i}}. \]
Support and target sets are set up to resemble a training and a validation set. Therefore the outer loop's objective becomes to \textit{maximize the performance on a validation set, given a training set}, hence pushing towards generalization (additional details are provided in Appendix \ref{episode_algo}).

\subsection{Model Architecture Design}\label{architecture}
We use an encoder-decoder model with a multi-head architecture. The \textit{backbone} (which represents the encoder) is a 3 layer GCN \citep{kipf2017semi}, while the decoder is composed of three \textit{heads} (one per task) with standard architectures. For additional information we refer the interested reader to Appendix \ref{arch}.

\subsection{Meta-Training Design}

To avoid the problems arising from training a model that performs multiple tasks concurrently, we design a meta-learning procedure where the inner loop adaptation and the meta-objective computation involve a \textit{single task} at a time. Only the parameter update performed to minimize the meta-objective involves multiple tasks, but,  crucially, it does not aim at a setting of parameters that can solve, or quickly adapt to, multiple tasks \textit{concurrently}, but to a setting that allows \textit{multiple} \textbf{fast single-task adaptation}. 
\begin{wrapfigure}{R}{0.57\textwidth}
   \begin{algorithm}[H]         
      \SetAlgoLined
      \DontPrintSemicolon
      \SetCustomAlgoRuledWidth{0.57\textwidth}
      \SetKwInOut{Input}{Input}
      \Input{Model $f_{\theta}$; Episodes $\mathcal{E} = \{ \mathcal{E}_1, .., \mathcal{E}_n \}$}     
      
      \texttt{init}$(\theta)$  \; 
      \For{$\mathcal{E}_i$ in $\mathcal{E}$}{
         $\text{\texttt{o\_loss}} \leftarrow 0$ \;
         \For{\text{\texttt{t}} in (GC, NC, LP)}{
            $\theta^{\prime (\text{\texttt{t}})} \leftarrow \theta$ \;
            $\theta^{\prime (\text{\texttt{t}})} \leftarrow \text{\texttt{ADAPT}}(f_{\theta}, \mathcal{S}^{(\texttt{t})}_{\mathcal{E}_i}, \mathcal{L}^{(\texttt{t})}_{\mathcal{E}_i})$ \;
            $\text{\texttt{o\_loss}} \leftarrow \text{\texttt{o\_loss}} + \text{\texttt{TEST}}(f_{\theta^{\prime (\text{\texttt{t}})}}, \mathcal{T}^{(\texttt{t})}_{\mathcal{E}_i}, \mathcal{L}^{(\texttt{t})}_{\mathcal{E}_i})$
         }
         $\theta \leftarrow \text{\texttt{UPDATE}}(\theta, \text{\texttt{o\_loss}}, \theta^{\prime (GC)}, \theta^{\prime (NC)}, \theta^{\prime (LP)})$
      }
       
       \caption{Proposed Meta-Learning Procedure}\label{meta_alg}
   \end{algorithm}
\end{wrapfigure}
The pseudocode of our procedure is in Algorithm \ref{meta_alg}. \texttt{ADAPT} performs a few steps of gradient descent on a task specific loss function and support set, \texttt{TEST} computes the value of a meta-objective component on a task specific loss function and target set, and \texttt{UPDATE} optimizes the parameters by minimizing the meta-objective. Notice how the multiple \textit{heads} of the decoder in our model are never used concurrently.

Let us partition the parameters $\theta$ of our model in four sets: one representing the backbone ($\theta_{GCN}$), and one for each head ($\theta_{NC},\theta_{GC},\theta_{LP}$).
We name our meta-learning strategy SAME (\underline{S}ingle-Task \underline{A}daptation for \underline{M}ulti-Task \underline{E}mbeddings), and present two variants (Figure \ref{fig:models} (b)-(c)): \textit{implicit} SAME (iSAME), and \textit{explicit} SAME (eSAME). In iSAME all the parameters $\theta$ are used for adaptation. iSAME makes use of the \textit{implicit} feature-reuse factor of MAML, leading to parameters $\theta_{\text{GCN}}$ that are general across \textit{multi-task episodes}. In eSAME only the head parameters $\theta_{\text{NC}}, \theta_{\text{GC}}, \theta_{\text{LP}}$ are used for adaptation. eSAME \textit{explicitly} aims at parameters $\theta_{\text{GCN}}$ that are general across \textit{multi-task episodes} by only updating them in the outer loop.


\section{Experiments}
Our goal is to assess the quality of the representations learned by our proposed method by answering four questions (\textbf{Q1}-\textbf{Q4}). 
Furthermore, by examining the results of the two variants of SAME, we can observe if the explicit strategy applied by eSAME is necessary for obtaining useful features, or if the implicit mechanism of iSAME is enough.
We use GC to refer to graph classification, NC for node classification, and LP for link prediction. Unless otherwise stated, accuracy (\%) is used for NC and GC, while ROC AUC (\%) is used for LP.
\begin{table}
  \caption{Results for a single-task model trained in a classical supervised manner (Cl), and a \textbf{linear} classifier trained on the embeddings produced by our meta-learning strategies (iSAME, eSAME).}\label{tableQ1} 
  \begin{center}
  \begin{tabular}{llcccc}
    \hline
    \textbf{Task} & \textbf{Model} & \multicolumn{4}{c}{\textbf{Dataset}} \\
            &            & ENZYMES & PROTEINS & DHFR & COX2 \\
    \hline
    \multirow{3}{*}{NC} & Cl & $87.5 \pm 1.9$ & $72.3 \pm 4.4$ & $97.3 \pm 0.2$ & $96.4 \pm 0.3$\\
                                    & iSAME & $87.3 \pm 0.8$ & $81.8 \pm 1.6$ & $96.6 \pm 0.3$ & $96.1 \pm 0.4$\\
                                    & eSAME & $87.8 \pm 0.7$ & $82.4 \pm 1.6$ & $96.8 \pm 0.2$ & $96.5 \pm 0.6$\\
    \hline
    \multirow{3}{*}{GC} & Cl & $51.6 \pm 4.2$ & $73.3 \pm 3.6$ & $71.5 \pm 2.3$ & $76.7 \pm 4.7$\\
                                   & iSAME & $50.8 \pm 2.9$ & $73.5 \pm 1.2$ & $73.2 \pm 3.2$ & $76.3 \pm 4.6$\\
                                   & eSAME & $52.1 \pm 5.0$ & $72.6 \pm 1.6$ & $71.6 \pm 2.4$ & $75.6 \pm 4.1$\\
    \hline
    \multirow{3}{*}{LP} & Cl & $75.5 \pm 3.0$ & $85.6 \pm 0.8$ & $98.8 \pm 0.7$ & $98.3 \pm 0.8$\\
                                   & iSAME& $81.7 \pm 1.7$ & $84.0 \pm 1.1$ & $99.2 \pm 0.4$ & $99.1 \pm 0.5$\\
                                   & eSAME & $80.1 \pm 3.4$ & $84.1 \pm 0.9$ & $99.2 \pm 0.3$ & $99.2 \pm 0.7$\\
    \hline
  \end{tabular}
  \end{center}
\end{table}

\textbf{Experimental Setup.} 
We consider datasets from the TUDataset library \citep{Morris+2020} that allow multi-task settings, and
perform a 10-fold cross validation. 
To ensure a fair comparison, we use the same architecture for all training strategies. 
For more information we refer to Appendix \ref{exp_details}.

\textbf{Q1: \textit{Do iSAME and eSAME lead to high quality node embeddings in the single-task setting?}}
For every task, we train a \textbf{linear classifier} on top of the embeddings produced by a model trained using our proposed methods, and compare against a network with the same architecture trained in a classical manner. Results are shown in Table \ref{tableQ1}. For all tasks, the \textbf{linear} classifier achieves comparable, if not superior, performance to the end-to-end model. In fact, the linear classifier is never outperformed by more than 2\%, and it can outperform the classical end-to-end model by up to 12\%. 

\begin{wrapfigure}{R}{0.52\textwidth}
\centering
  \subfigure[]{\includegraphics[width=0.165\textwidth]{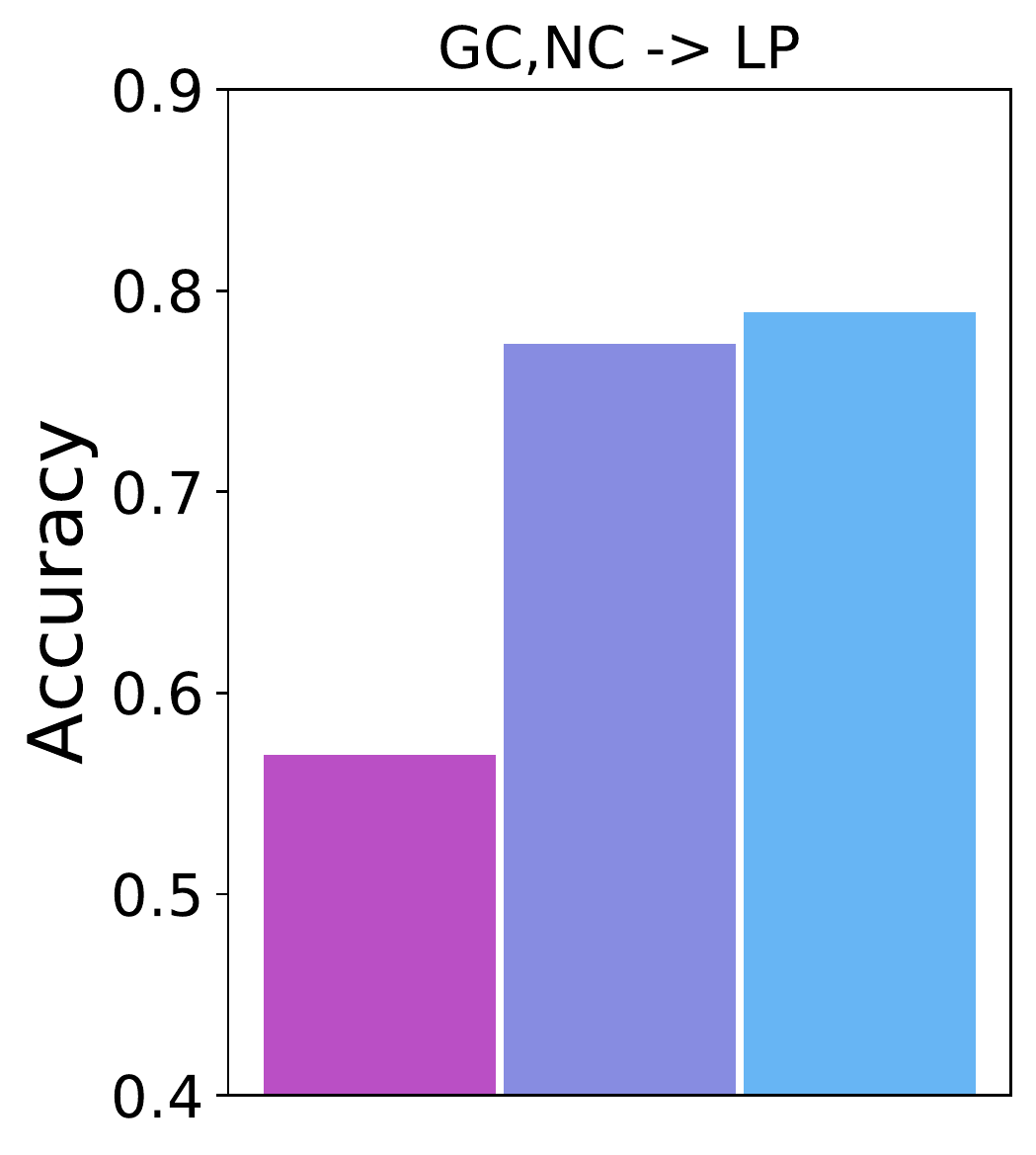}}
  \subfigure[]{\includegraphics[width=0.165\textwidth]{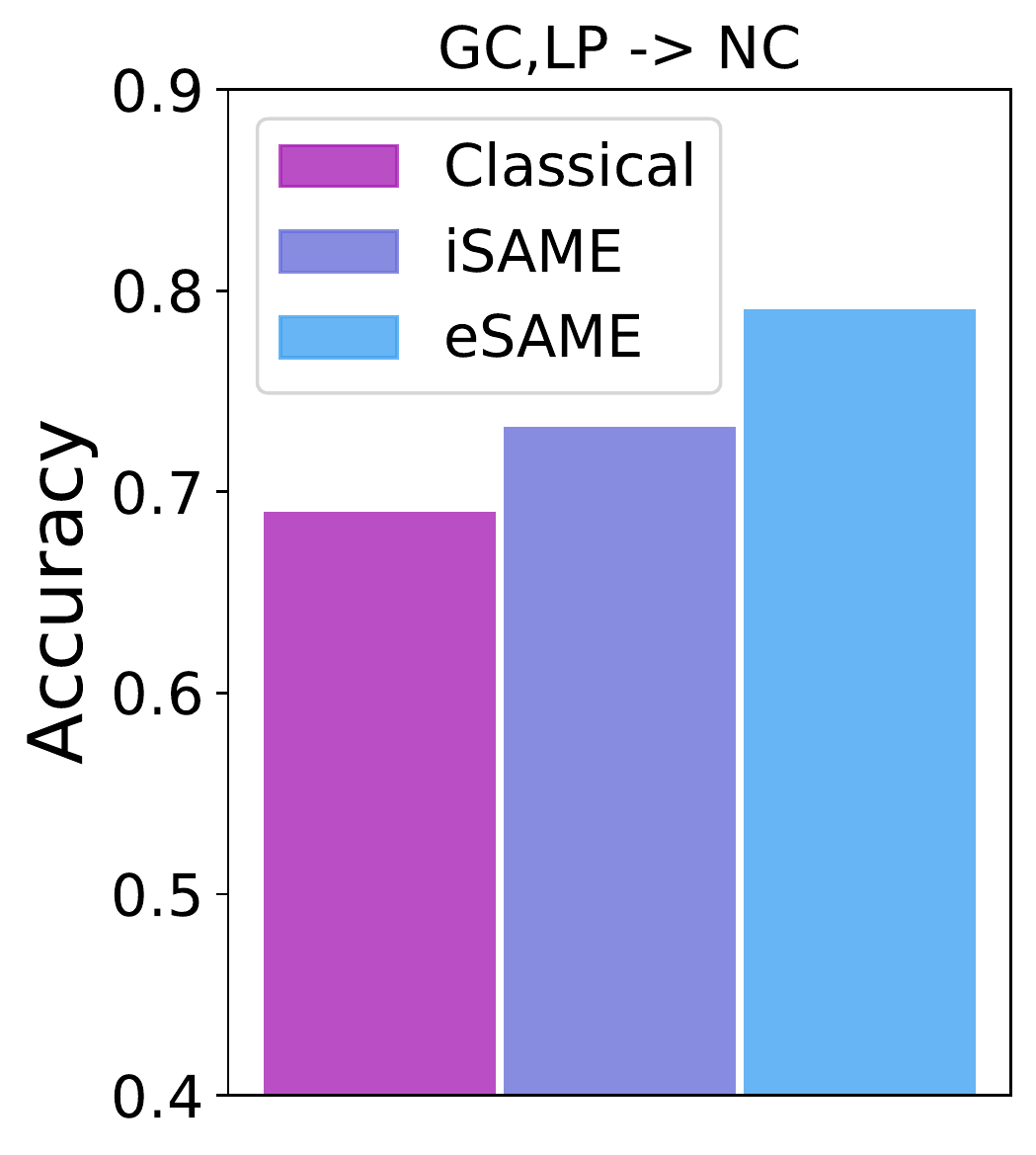}}
  \subfigure[]{\includegraphics[width=0.165\textwidth]{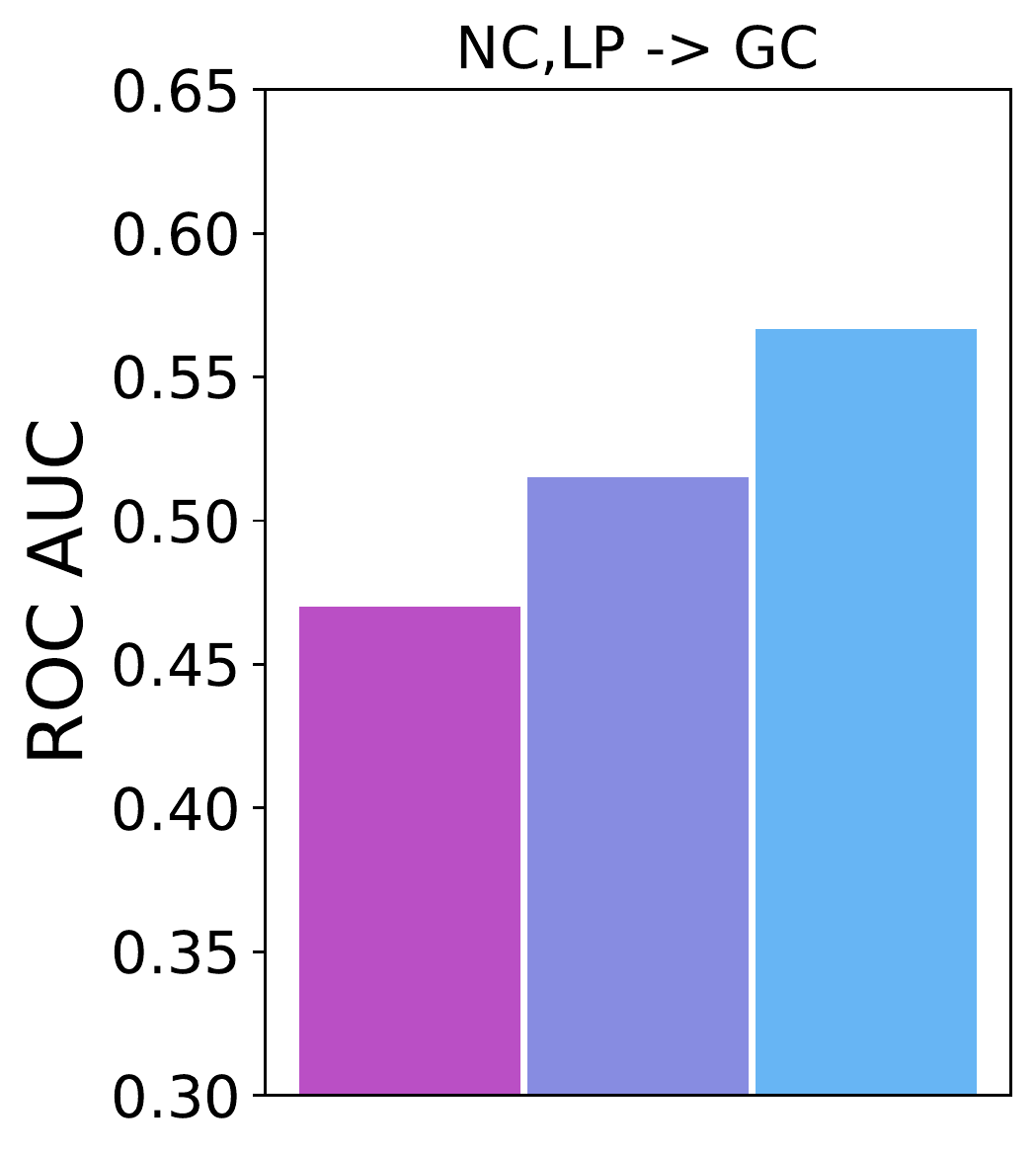}}
  \caption{Results for neural network, trained on the embeddings generated by a multi-task model, performing a task that was not seen by the multi-task model. ``$x,y$-\textgreater$z$'' indicates that $x,y$ are the tasks for training the multi-task model, and $z$ is the new task.}
  \label{fig:transfer_multitask_embeddings}
\end{wrapfigure}
\textbf{Q2: \textit{Do iSAME and eSAME lead to high quality node embeddings in the multi-task setting?}}
We train a model with our proposed methods, on all multi-task combinations, and use the embeddings as input for a \textbf{linear classifier}. We compare against single-task models trained in the classical manner, and with a fine-tuning baseline. The latter is a model that has been trained on all three tasks, and then fine-tuned on two specific tasks. The idea is that the initial training on all tasks should lead the model towards the extraction of features that it would otherwise not consider (by only seeing 2 tasks), and the fine-tuning process should then allow the model to use these features to target the specific tasks of interest. Results are shown in Table \ref{tableQ2}. We notice that the \textbf{linear} classifier, achieves comparable performance to the end-to-end models, as it is never outperformed by more than 3\%, and in 50\% of the cases it actually performs better, confirming the high quality of the node embeddings learned with iSAME and eSAME. We further notice that the fine-tuning baseline severely struggles, and is almost always outperformed by both single-task models, and our proposed methods. These results indicate that the episodic meta-learning procedure adopted by SAME is extracting features that are otherwise not accessible with standard training techniques.

\begin{table}
  \caption{Results for a single-task model trained in a classical supervised manner, a fine-tuned model (trained on all three tasks, and fine-tuned on the two shown tasks), and a \textbf{linear} classifier trained on node embeddings learned with our proposed strategies (iSAME, eSAME) in a multi-task setting.}\label{tableQ2}
  \begin{center}
  \begin{tabular}{ccccccc}
    \hline
    \multicolumn{3}{c}{\textbf{Task}} & \multicolumn{4}{c}{\textbf{Dataset}} \\
                GC & NC & LP                 & ENZYMES & PROTEINS & DHFR & COX2 \\
               &&& GC \phantom{ /} NC \phantom{/ } LP & GC \phantom{ /} NC \phantom{/ } LP & GC \phantom{ /} NC \phantom{/ } LP & GC \phantom{ /} NC \phantom{/ } LP\\
    \hline
    \multicolumn{7}{c}{\textbf{Classical End-to-End Training}}\\
    \hline
     \checkmark &  &                    & 51.6 \phantom{/} \phantom{00.0} \phantom{/} \phantom{00.0} & 73.3 \phantom{/} \phantom{00.0} \phantom{/} \phantom{00.0} & 71.5 \phantom{/} \phantom{00.0} \phantom{/} \phantom{00.0} & 76.7 \phantom{/} \phantom{00.0} \phantom{/} \phantom{00.0}\\
                        & \checkmark &                                & \phantom{00.0} \phantom{/} 87.5 \phantom{/} \phantom{00.0} & \phantom{00.0} \phantom{/} 72.3 \phantom{/} \phantom{00.0} & \phantom{00.0} \phantom{/} 97.3 \phantom{/} \phantom{00.0} & \phantom{00.0} \phantom{/} 96.4 \phantom{/} \phantom{00.0}\\
                        &  & \checkmark                                & \phantom{00.0} \phantom{/} \phantom{00.0} \phantom{/} 75.5 & \phantom{00.0} \phantom{/} \phantom{00.0} \phantom{/} 85.6 & \phantom{00.0} \phantom{/} \phantom{00.0} \phantom{/} 98.8 & \phantom{00.0} \phantom{/} \phantom{00.0} \phantom{/} 98.3\\
    \hline
    \multicolumn{7}{c}{\textbf{Fine-Tuning}}\\
    \hline
    \checkmark & \checkmark &                     & 48.3 \phantom{/} 85.3 \phantom{/} \phantom{00.0} & 73.6 \phantom{/} 72.0 \phantom{/} \phantom{00.0} & 66.4 \phantom{/} 92.4 \phantom{/} \phantom{00.0} & 80.0 \phantom{/} 92.3 \phantom{/} \phantom{00.0}\\
     \checkmark &  & \checkmark                                                       & 49.3 \phantom{/} \phantom{00.0} \phantom{/} 71.6 & 69.6 \phantom{/} \phantom{00.0} \phantom{/} 80.7 & 65.3 \phantom{/} \phantom{00.0} \phantom{/} 58.9 & 80.2 \phantom{/} \phantom{00.0} \phantom{/} 50.9\\
     & \checkmark  & \checkmark                                                       & \phantom{00.0} \phantom{/} 87.7 \phantom{/} 73.9 & \phantom{00.0} \phantom{/} 80.4 \phantom{/} 81.5 & \phantom{00.0} \phantom{/} 80.7 \phantom{/} 56.6 & \phantom{00.0} \phantom{/} 87.4 \phantom{/} 52.3\\
    \hline
    \multicolumn{7}{c}{\textbf{iSAME (ours)}}\\
    \hline
     \checkmark & \checkmark &                     & 50.1 \phantom{/} 86.1 \phantom{/} \phantom{00.0} & 73.1 \phantom{/} 76.6 \phantom{/} \phantom{00.0} & 71.6 \phantom{/} 94.8 \phantom{/} \phantom{00.0} & 75.2 \phantom{/} 95.4 \phantom{/} \phantom{00.0}\\
     \checkmark &  & \checkmark                                                       & 50.7 \phantom{/} \phantom{00.0} \phantom{/} 83.1 & 73.4 \phantom{/} \phantom{00.0} \phantom{/} 85.2 & 71.6 \phantom{/} \phantom{00.0} \phantom{/} 99.2 & 77.5 \phantom{/} \phantom{00.0} \phantom{/} 98.9\\
     & \checkmark  & \checkmark                                                       & \phantom{00.0} \phantom{/} 86.3 \phantom{/} 83.4 & \phantom{00.0} \phantom{/} 79.4 \phantom{/} 87.7 & \phantom{00.0} \phantom{/} 96.5 \phantom{/} 99.3 & \phantom{00.0} \phantom{/} 95.5 \phantom{/} 99.0\\
     \checkmark & \checkmark  & \checkmark                                    & 50.0 \phantom{/} 86.5 \phantom{/} 82.3 & 71.4 \phantom{/} 76.6 \phantom{/} 87.3 & 71.2 \phantom{/} 95.5 \phantom{/} 99.5 & 75.4 \phantom{/} 95.2 \phantom{/} 99.2\\
    \hline
    \multicolumn{7}{c}{\textbf{eSAME (ours)}}\\
    \hline
     \checkmark & \checkmark &                     & 51.7 \phantom{/} 86.1 \phantom{/} \phantom{00.0} & 71.5 \phantom{/} 79.2 \phantom{/} \phantom{00.0} & 70.1 \phantom{/} 95.7 \phantom{/} \phantom{00.0} & 75.6 \phantom{/} 95.5 \phantom{/} \phantom{00.0}\\
     \checkmark &  & \checkmark                                                       & 51.9 \phantom{/} \phantom{00.0} \phantom{/} 80.1 & 71.7 \phantom{/} \phantom{00.0} \phantom{/} 85.4 & 70.1 \phantom{/} \phantom{00.0} \phantom{/} 99.1 & 77.5 \phantom{/} \phantom{00.0} \phantom{/} 98.8\\
     & \checkmark  & \checkmark                                                       & \phantom{00.0} \phantom{/} 86.7 \phantom{/} 82.2 & \phantom{00.0} \phantom{/} 80.7 \phantom{/} 86.3 & \phantom{00.0} \phantom{/} 96.6 \phantom{/} 99.4 & \phantom{00.0} \phantom{/} 95.6 \phantom{/} 99.1\\
     \checkmark & \checkmark  & \checkmark                                    & 51.5 \phantom{/} 86.3 \phantom{/} 81.1 & 71.3 \phantom{/} 79.6 \phantom{/} 86.8 & 70.2 \phantom{/} 95.3 \phantom{/} 99.5 & 77.7 \phantom{/} 95.7 \phantom{/} 98.8\\
    \hline
  \end{tabular}
  \end{center}
\end{table}

\textbf{Q3: \textit{Do iSAME and eSAME extract information that is not captured by classically trained multi-task models?}}
We train a network, which we refer to as \textit{classifier}, on the embeddings generated by a multi-task model, to perform a task that was not seen during the training of the latter.
We compare the performance of the classifier on the embeddings learned by a model trained in a classical manner, and with our proposed methods. 
This test allows us to quantify if our approaches lead to ``more informative'' node embeddings.
Results on the ENZYMES dataset are shown in Figure \ref{fig:transfer_multitask_embeddings}. We notice that embeddings learned by our proposed approaches lead to at least 10\% higher performance. We observe an analogous trend on the other datasets (as reported in Appendix \ref{full_res_q3}).

\begin{table}[t]
  \caption{$\Delta_m$ (\%) results for a classical multi-task model (Cl), a fine-tuned model (FT; trained on all three tasks and fine-tuned on two) and a \textbf{linear} classifier trained on the node embeddings learned using our meta-learning strategies (iSAME, eSAME) in a multi-task setting.}\label{tableQ3}
  \begin{center}
  \begin{tabular}{cccccccc}
    \hline
    \multicolumn{3}{c}{\textbf{Task}} & \textbf{Model} & \multicolumn{4}{c}{\textbf{Dataset}} \\
               GC & NC & LP     &            & ENZYMES & PROTEINS & DHFR & COX2 \\
    \hline
    \multirow{4}{*}{\checkmark} & \multirow{4}{*}{\checkmark} & \multirow{4}{*}{ } & Cl & $-0.1 \pm 0.5$ & $4.0 \pm 1.0$ & $-0.3 \pm 0.2$ & $0.5 \pm 0.1$\\
                                                     &&                                                                    & FT & $-4.5 \pm 1.2$ & $0.1 \pm 0.5$ & $-7.4 \pm 1.4$ & $0.1 \pm 0.4$\\
                                                     &&                                                                    & iSAME & $-2.3 \pm 0.9$ & $2.7 \pm 1.5$ & $-1.2 \pm 0.4$ & $-1.6 \pm 0.2$\\
                                                     &&                                                                    & eSAME & $-0.8 \pm 0.8$ & $3.2 \pm 1.4$ & $-1.8 \pm 0.3$ & $-1.2 \pm 0.3$\\
    \hline
    \multirow{4}{*}{\checkmark} & \multirow{4}{*}{ } & \multirow{4}{*}{\checkmark} & Cl & $-25.3 \pm 3.2$ & $-5.3 \pm 1.2$ & $-28.3 \pm 4.3$ & $-21.4 \pm 3.4$\\
                                                         &&                                                                    & FT & $-5.1 \pm 1.9$ & $-5.4 \pm 1.5$ & $-24.5 \pm 3.7$ & $-22.6 \pm 3.8$\\
                                                     &&                                                                    & iSAME & $4.1 \pm 0.5$ & $-0.2 \pm 0.9$ & $0.2 \pm 3.2$ & $0.2 \pm 0.5$\\
                                                     &&                                                                    & eSAME & $3.2 \pm 0.4$ & $-1.2 \pm 1.1$ & $-0.7 \pm 3.4$ & $-0.8 \pm 0.7$\\
    \hline
    \multirow{4}{*}{ } & \multirow{4}{*}{\checkmark} & \multirow{4}{*}{\checkmark} & Cl & $7.2 \pm 2.7$ & $6.8 \pm 0.9$ & $-29.1 \pm 7.7$ & $-28.2 \pm 4.5$\\
                                                         &&                                                                    & FT & $-1.0 \pm 0.3$ & $3.1 \pm 1.2$ & $-28.9 \pm 6.4$ & $-28.3 \pm 4.2$\\
                                                     &&                                                                    & iSAME & $4.4 \pm 1.1$ & $6.1 \pm 1.0$ & $-0.1 \pm 6.2$ & $-0.6 \pm 2.5$\\
                                                     &&                                                                    & eSAME & $3.9 \pm 1.3$ & $6.1 \pm 1.1$ & $0.1 \pm 6.4$ & $-0.6 \pm 2.6$\\
    \hline
    \multirow{3}{*}{\checkmark} & \multirow{3}{*}{\checkmark} & \multirow{3}{*}{\checkmark} & Cl & $ 1.6 \pm 1.3$ & $2.9 \pm 0.3$ & $-18.9 \pm 2.3$ & $-16.9 \pm 3.1$\\
                                                     &&                                                                    & iSAME & $1.5 \pm 1.0$ & $2.2 \pm 0.2$ & $-0.5 \pm 1.4$ & $-0.9 \pm 1.3$\\
                                                     &&                                                                    & eSAME & $1.8 \pm 0.9$ & $2.8 \pm 0.2$ & $-1.0 \pm 1.7$ & $-0.4 \pm 1.2$\\
    \hline
  \end{tabular}
  \end{center}
\end{table}

\textbf{Q4: \textit{Can the node embeddings learned by iSAME and eSAME be used to perform multiple tasks with comparable or better performance than classical multi-task models?}} 
We train the same multi-task model, both in the classical supervised manner, and with our proposed approaches, on all multi-task combinations. For our approaches, we then train a \textbf{linear classifier} on top of the node embeddings. We further consider the fine-tuning baseline introduced in \textbf{Q2}. We use the $\Delta_m$ metric \citep{8954118} , defined as the average per-task drop with respect to the single-task baseline:
$\Delta_m = \frac{1}{T} \sum_{i=1}^{T} \left(M_{m,i} - M_{b,i} \right) / M_{b,i},$
where $M_{m,i}$ is the value of a task's metric for the multi-task model, and $M_{b,i}$ is the value for the baseline. 
Results are shown in Table \ref{tableQ3}. We first notice that usually multi-task models achieve lower performance than specialized single-task ones. We then highlight that \textbf{linear} classifiers trained on the embeddings produced by our procedures are comparable, and in many cases superior, to end-to-end models. In fact, the latter are highly sensible to the tasks that are being learned (e.g. GC and LP), with a worst-case average drop in performance of 29\%. Our methods seem much less sensible, with a worst-case average drop of less than 3\%. Finally, we also notice that the fine-tuning baseline generally performs worst than classically trained models, confirming that transferring knowledge in multi-task settings is not easy, and more advanced techniques, like our proposed method SAME, are needed.

\textbf{Considerations on iSAME and eSAME.} In all our experiments we notice that the performance between the two variants of SAME achieve comparable results. This suggests that the representation learning capabilities are an intrinsic property of optimization-based meta-learning approaches like MAML \citep{finn2017}, and that strategies like ANIL \citep{raghu2019rapid} can help us lower the computational burden, while maintaining the desired properties.

\section{Conclusions}\label{conclusions}
We propose a novel meta-learning strategy for representation learning in multi-task settings.
Our method overcomes the problems that arise when learning to solve multiple tasks concurrently by optimizing for a parameter setting that can quickly, i.e. with few steps of gradient descent, be adapted for high \textit{single-task} performance on multiple tasks.
We apply our method to graph representation learning, and find that it leads to higher quality node embeddings, both in the multi-task and in the single-task setting. 
We believe this work draws new interesting connections between meta-learning, representation learning, and multi-task learning, providing many directions for future research.

\begin{ack}
Part of this work was supported by the MIUR, the Italian Ministry of Education, University and Research, under PRIN Project n. 20174LF3T8 AHeAD (Efficient Algorithms for HArnessing Networked Data) and the initiative “Departments of Excellence” (Law 232/2016), and by the University of Padova under project SEED 2020 RATED-X.
\end{ack}

\bibliography{iclr2021_conference}
\bibliographystyle{plainnat}

\appendix
\section{Comparison with Traditional Training Apporaches}\label{appendix_diff}
Our proposed meta-learning approach is significantly different from the classical training strategy (Algorithm \ref{classic_alg}), and the traditional meta-learning approaches (Algorithm \ref{trad_meta_alg}). 

The classical training approach for multi-task models takes as input a \textit{batch} of graphs, which is simply a set of graphs, where on each graph the model has to execute \textit{all} the tasks. Based on the cumulative loss on all tasks 
\[ \mathcal{L} = \lambda^{(GC)} \mathcal{L}^{(\text{GC})} + \lambda^{(NC)} \mathcal{L}^{(\text{NC})} + \lambda^{(LP)} \mathcal{L}^{(\text{LP})} \] 
for all the graphs in the batch, the parameters are updated with some form of gradient descent, and the procedure is repeated for each batch.

The traditional meta-learning approach takes as input an episode, like our approach, but for every graph in the episode \textit{all} the tasks are performed. The support set and target set are \textit{single} sets of graphs, where every task can be performed on all graphs. The support set is used to obtain the adapted parameters $\theta^{\prime}$, which have the goal of \textit{concurrently} solving all tasks on all graphs in the target set. The loss functions, both for the inner loop and for the outer loop, are the same as the one used by the classical training approach. The outer loop then updates the parameters aiming at a setting that can easily, i.e. with a few steps of gradient descent, be adapted to perform multiple tasks \textit{concurrently} given a support set.

 \begin{minipage}[t]{0.495\textwidth}
  \begin{algorithm}[H]         
      \SetAlgoLined
      \DontPrintSemicolon
      \SetCustomAlgoRuledWidth{0.49\textwidth}
      \SetKwInOut{Input}{Input}
      \Input{Model $f_{\theta}$; Batches $\mathcal{B} = \{ \mathcal{B}_1, .., \mathcal{B}_n \}$}     
      
      \texttt{init}$(\theta)$  \;
      \For{$\mathcal{B}_i$ in $\mathcal{B}$}{
         $\text{\texttt{loss}} \leftarrow$ concurrently perform all tasks on all graphs in $\mathcal{B}_i$ \;
         $\theta \leftarrow \text{\texttt{UPDATE}}(\theta, \text{\texttt{loss}})$
      }
       
       \caption{Classical Training}\label{classic_alg}
   \end{algorithm}
 \end{minipage}\hfill
 \begin{minipage}[t]{0.495\textwidth}
  \begin{algorithm}[H]         
      \SetAlgoLined
      \DontPrintSemicolon
      \SetCustomAlgoRuledWidth{0.49\textwidth}
      \SetKwInOut{Input}{Input}
      \Input{Model $f_{\theta}$; Episodes $\mathcal{E} = \{ \mathcal{E}_1, .., \mathcal{E}_n \}$}     
       
      \texttt{init}$(\theta)$  \;
      \For{$\mathcal{E}_i$ in $\mathcal{E}$}{
         $\text{\texttt{i\_loss}} \leftarrow$ concurrently perform all tasks on all support set graphs \;
         $\theta^{\prime} \leftarrow \text{\texttt{ADAPT}}(\theta, \text{\texttt{i\_loss}} )$ \;
         $\text{\texttt{o\_loss}} \leftarrow$ concurrently perform all tasks on all target set graphs using parameters $\theta^{\prime}$ \;
         $\theta \leftarrow \text{\texttt{UPDATE}}(\theta, \theta^{\prime}, \text{\texttt{o\_loss}})$
      }
       
       \caption{Traditional Meta-Learning}\label{trad_meta_alg}
   \end{algorithm}
 \end{minipage}

\section{Episode Design Algorithm}\label{episode_algo}
Algorithm \ref{epi_alg} contains the procedure for the creation of the episodes for our meta-learning procedures. The algorithm takes as input a batch of graphs (with graph labels, node labels, and node features) and the loss function balancing weights, and outputs a \textit{multi-task episode}. We assume that each graph has a set of attributes that can be accessed with a \textit{dot-notation} (like in most object-oriented programming languages).

Notice how the episodes are created so that only one task is performed on each graph. This is important as in the inner loop of our meta-learning procedure, the learner adapts and tests the adaptated parameters on one task at a time. The outer loop then updates the parameters, optimizing for a representation that leads to fast \textit{single-task adaptation}. This procedure bypasses the problem of learning parameters that \textit{directly} solve multiple tasks, which can be very challenging. 

Another important aspect to notice is that the support and target sets are designed as if they were the training and validation splits for training a single-task model with the classical procedure. This way the meta-objective becomes to train a model that can generalize well. 

\begin{algorithm}
\SetAlgoLined
\DontPrintSemicolon
\SetKwInOut{Input}{Input}
\SetKwInOut{Output}{Output}
\Input{Batch of $n$ randomly sampled graphs $\mathcal{B} = \{ \mathcal{G}_1, .., \mathcal{G}_n \}$ \\ Loss weights $\lambda^{(GC)}, \lambda^{(NC)}, \lambda^{(LP)} \in [0,1]$}
\Output{Episode $\mathcal{E}_i = ( \mathcal{L}^{(m)}_{\mathcal{E}_{i}},  \mathcal{S}^{(m)}_{\mathcal{E}_{i}},   \mathcal{T}^{(m)}_{\mathcal{E}_{i}} )$} \;
$\mathcal{B}^{(GC)}, \mathcal{B}^{(NC)}, \mathcal{B}^{(LP)} \leftarrow$ equally divide the graphs in $\mathcal{B}$ in three sets \; \;
 \tcc{Graph Classification}
  $\mathcal{S}^{(\text{GC})}_{\mathcal{E}_{i}}, \mathcal{T}^{(\text{GC})}_{\mathcal{E}_{i}} \leftarrow$ randomly divide $\mathcal{B}^{(GC)}$ with a 60/40 split\; \;
  
  \tcc{Node Classification}
  \For{$\mathcal{G}_i$ in $\mathcal{B}^{(NC)}$}{
  $\text{\texttt{num\_labelled\_nodes}} \leftarrow \mathcal{G}_{i}\text{\texttt{.num\_nodes}} \times  0.3$ \;
  $\mathcal{N} \leftarrow$ divide nodes per class, then iteratively randomly sample one node per class without replacement and add it to $\mathcal{N}$ until $\lvert \mathcal{N} \vert = \text{\texttt{num\_labelled\_nodes}}$ \;
  $\mathcal{G}_{i}^{\prime} \leftarrow \text{\texttt{copy}}(\mathcal{G}_i)$ \;
  $\mathcal{G}_{i}\text{\texttt{.labelled\_nodes}} \leftarrow \mathcal{N}$; $\quad \mathcal{G}_{i}^{\prime}\text{\texttt{.labelled\_nodes}} \leftarrow \mathcal{G}_{i}\text{\texttt{.nodes}} \setminus \mathcal{N}$ \;
  $\mathcal{S}^{(NC)}_{\mathcal{E}_{i}}\text{\texttt{.add}}(\mathcal{G}_{i})$; $\quad \mathcal{T}^{(NC)}_{\mathcal{E}_{i}}\text{\texttt{.add}}(\mathcal{G}_{i}^{\prime})$
 }
 
 \;
 \tcc{Link Prediction}
  \For{$\mathcal{G}_i$ in $\mathcal{B}^{(LP)}$}{
  $E_{i}^{(N)} \leftarrow$ randomly pick negative samples (edges that are not in the graph; possibly in the same number as the number of edges in the graph) \;
  $E_{i}^{1,(N)}, E_{i}^{2,(N)} \leftarrow$ divide $E_{i}^{(N)}$ with an $80/20$ split \;
  $E_{i}^{(P)} \leftarrow$ randomly remove $20\%$ of the edges in $\mathcal{G}_{i}$ \;
  $\mathcal{G}_{i}^{\prime (1)} \leftarrow \mathcal{G}_{i}$ removed of $E_{i}^{(P)}$ \;
  $\mathcal{G}_{i}^{\prime (2)} \leftarrow \text{\texttt{copy}}( \mathcal{G}_{i}^{\prime (1)})$ \;
  $\mathcal{G}_{i}^{\prime (1)} \text{\texttt{.positive\_edges}} \leftarrow \mathcal{G}_{i}^{\prime (1)} \text{\texttt{.edges}}$; $\quad \mathcal{G}_{i}^{\prime (2)} \text{\texttt{.positive\_edges}} \leftarrow E_{i}^{(P)} $ \;
  $\mathcal{G}_{i}^{\prime (1)} \text{\texttt{.negative\_edges}} \leftarrow E_{i}^{1,(N)}$; $\quad \mathcal{G}_{i}^{\prime (2)} \text{\texttt{.negative\_edges}} \leftarrow E_{i}^{2,(N)}$ \;
  $\mathcal{S}^{(LP)}_{\mathcal{E}_{i}}\text{\texttt{.add}}(\mathcal{G}_{i}^{\prime (1)})$; $\quad \mathcal{T}^{(LP)}_{\mathcal{E}_{i}}\text{\texttt{.add}}(\mathcal{G}_{i}^{\prime (2)})$
 }
 \;
 $\mathcal{S}^{(m)}_{\mathcal{E}_{i}} \leftarrow \{ \mathcal{S}^{(\text{GC})}_{\mathcal{E}_{i}}, \mathcal{S}^{(\text{NC})}_{\mathcal{E}_{i}}, \mathcal{S}^{(\text{LP})}_{\mathcal{E}_{i}} \}$ \;
  $\mathcal{T}^{(m)}_{\mathcal{E}_{i}} \leftarrow \{ \mathcal{T}^{(\text{GC})}_{\mathcal{E}_{i}}, \mathcal{T}^{(\text{NC})}_{\mathcal{E}_{i}}, \mathcal{T}^{(\text{LP})}_{\mathcal{E}_{i}} \}$ \;
 $\mathcal{L}^{(\text{GC})}_{\mathcal{T}_{i}} \leftarrow $ \texttt{Cross-Entropy}$(\cdot)$; $\quad \mathcal{L}^{(\text{NC})}_{\mathcal{T}_{i}} \leftarrow $ \texttt{Cross-Entropy}$(\cdot)$ \;
 $\mathcal{L}^{(\text{LP})}_{\mathcal{T}_{i}} \leftarrow $ \texttt{Binary Cross-Entropy}$(\cdot)$ \;
 $ \mathcal{L}^{(m)}_{\mathcal{E}_{i}} = \lambda^{(GC)} \mathcal{L}^{(\text{GC})}_{\mathcal{T}_{i}} + \lambda^{(NC)} \mathcal{L}^{(\text{NC})}_{\mathcal{T}_{i}} + \lambda^{(LP)} \mathcal{L}^{(\text{LP})}_{\mathcal{T}_{i}}$ \;
 \textbf{Return} $\mathcal{E} = ( \mathcal{L}^{(m)}_{\mathcal{E}_{i}},  \mathcal{S}^{(m)}_{\mathcal{E}_{i}},   \mathcal{T}^{(m)}_{\mathcal{E}_{i}} )$
 \caption{Episode Design Algorithm}\label{epi_alg}
\end{algorithm}

\section{Model Architecture}\label{arch}
We use an encoder-decoder model with a multi-head architecture. The \textit{backbone} (which represents the encoder) is composed of 3 GCN \citep{kipf2017semi} layers with ReLU non-linearities and residual connections \citep{7780459}. The decoder is composed of three \textit{heads}. The node classification head is a single layer neural network with a \textit{Softmax} activation that is shared across nodes and maps node embeddings to class predictions. In the graph classification head, first a single layer neural network (shared across nodes) performs a linear transformation (followed by a ReLU activation) of the node embeddings. The transformed node embeddings are then averaged and a final single layer neural network with \textit{Softmax} activation outputs the class predictions. The link prediction head is composed of a single layer neural network with a ReLU non-linearity that transforms node embeddings, and another single layer neural network that takes as input the concatenation of two node embeddings and outputs the probability of a link between them.

\section{Additional Experimental Details}\label{exp_details}
In this section we provide additional information on the implementation of the models used in our experimental section. We implement our models using PyTorch \citep{NEURIPS2019_9015}, PyTorch Geometric \citep{Fey/Lenssen/2019} and Torchmeta \citep{deleu2019torchmeta}. For all models the number and structure of the layers is as described in Appendix \ref{arch}, where we use 256-dimensional node embeddings at every layer.

To perform multiple tasks, we consider datasets with graph labels, node attributes, and node labels from the widely used TUDataset library \citep{Morris+2020}. At every cross-validation fold the datasets are split into $70\%$ for training, $10\%$ for validation, and $20\%$ for testing. For each model we perform 100 iterations of hyperparameter optimization over the same search space (for shared parameters) using Ax \citep{Bakshy2018AEA}. 

We tried some sophisticated methods to balance the contribution of loss functions during multi-task training like GradNorm \citep{chen2018gradnorm} and Uncertainty Weights \citep{8578879}, but we saw that usually they do not positively impact performance. Furthermore, in the few cases where they increase performance, they work for both classically trained models, and for models trained with our proposed procedures. We then set the balancing weights to $\lambda^{(GC)} = \lambda^{(NC)} = \lambda^{(LP)} = 1$ to provide better comparisons between the training strategies. 

The multi-task performance $\Delta_m$ metric \citep{8954118} is defined as the average per-task drop with respect to the single-task baseline:
$\Delta_m = \frac{1}{T} \sum_{i=1}^{T} \left(M_{m,i} - M_{b,i} \right) / M_{b,i},$ where $M_{m,i}$ is the value for the multi-task model, and $M_{b,i}$ for the baseline. 

\paragraph{Linear Model.} The linear model trained on the embeddings produced by our proposed method is a standard linear SVM. In particular we use the implementation available in Scikit-learn \citep{scikit-learn} with default hyperparameters. For graph classification, we take the mean of the node embeddings as input. For link prediction we take the concatenation of the embeddings of two nodes. For node classification we keep the embeddings unaltered.
\paragraph{Deep Learning Baselines.} We train the single task models for 1000 epochs, and the multi-task models for 5000 epochs, with early stopping on the validation set (for multi-task models we use the sum of the task validation losses or accuracies as metrics for early-stopping). Optimization is done using Adam \citep{Kingma2015AdamAM}. For node classification and link prediction we found that normalizing the node embeddings to unit norm in between GCN layers helps performance.
\paragraph{Our Meta-Learning Procedure.} We train the single task models for 5000 epochs, and the multi-task models for 15000 epochs, with early stopping on the validation set (for multi-task models we use the sum of the task validation losses or accuracies as metrics for early-stopping). Early stopping is very important in this case as it is the only way to check if the meta-learned model is overfitting the training data.
The inner loop adaptation consists of 1 step of gradient descent. Optimization in the outer loop is done using Adam \citep{Kingma2015AdamAM}. We found that normalizing the node embeddings to unit norm in between GCN layers helps performance.

 \begin{table}[t]
  \caption{Results of a neural network trained on the embeddings generated by a multi-task model, to perform a task that was not seen during training by the multi-task model. ``$x$,$y$ -\textgreater $z$'' indicates that the multi-task model was trained on tasks $x$ and $y$, and the neural network is performing task $z$.}\label{tableQ3}
  \begin{center}
  \begin{tabular}{llcccc}
    \hline
    \textbf{Task} & \textbf{Model} & \multicolumn{4}{c}{\textbf{Dataset}} \\
            &            & ENZYMES & PROTEINS & DHFR & COX2 \\
    \hline
    \multirow{3}{*}{GC,NC -\textgreater LP} & Cl & $56.9 \pm 3.9$ & $54.4 \pm 1.4$ & $61.2 \pm 2.2$ & $59.8 \pm 0.4$\\
                                                                   & iSAME & $77.3 \pm 4.5$ & $88.5 \pm 1.8$ & $99.8 \pm 1.8$ & $97.1 \pm 2.0$\\
                                                                   & eSAME & $78.9 \pm 2.8$ & $89.1 \pm 1.5$ & $99.7 \pm 2.2$ & $95.8 \pm 3.3$\\
    \hline
    \multirow{3}{*}{GC,LP -\textgreater NC} & Cl & $69.1 \pm 1.2$ & $57.3 \pm 1.6$ & $58.3 \pm 9.3$ & $68.9 \pm 10.7$\\
                                                                   & iSAME & $73.3 \pm 2.1$ & $59.2 \pm 2.5$ & $77.6 \pm 1.6$ & $78.1 \pm 4.6$\\
                                                                   & eSAME & $79.1 \pm 1.7$ & $64.7 \pm 3.0$ & $76.1 \pm 2.7$ & $76.9 \pm 3.3$\\
    \hline
    \multirow{3}{*}{NC,LP -\textgreater GC} & Cl & $47.1 \pm 2.4$ & $75.3 \pm 1.5$ & $77.5 \pm 3.1$ & $79.9 \pm 3.4$\\
                                                                   & iSAME & $48.5 \pm 5.5$ & $76.1 \pm 2.3$ & $76.1 \pm 3.7$ & $79.7 \pm 5.1$\\
                                                                   & eSAME & $56.6 \pm 3.1$ & $74.6 \pm 2.7$ & $77.1 \pm 3.6$ & $79.3 \pm 6.2$\\
    \hline
  \end{tabular}
  \end{center}
\end{table}
 \section{Full Results for Q3}\label{full_res_q3}
 Table \ref{tableQ3} contains results for a neural network, trained on the embeddings generated by a multi-task model, to perform a task that was not seen during the training of the multi-task model.
 Accuracy (\%) is used for node classification (NC) and graph classification (GC); ROC AUC (\%) is used for link prediction (LP). The embeddings produced by our meta-learning methods lead to higher performance (up to \textbf{35\%}), showing that our procedures lead to the extraction of more informative node embeddings with respect to the classical end-to-end training procedure.

\end{document}